\documentclass[11pt]{article}

\newcommand{\dfrac}{\displaystyle\frac}

\newcommand{\bm}[1]{\mbox{\boldmath $#1$}}

\newcounter{alphnum}

\newcounter{Romnum}

\newcounter{romnum}

\newcounter{romnum2}

\newcounter{Romnum2}

\newtheorem{Corollary}{\bf Corollary}

\newtheorem{Example}{\bf Example}
\newtheorem{Proposition}{Proposition}
\newtheorem{Remark}{\bf Remark}
\usepackage{pgf,pgfarrows,pgfnodes,pgfautomata,pgfheaps,pgfshade}
\usepackage{tikz}
\usepackage{graphicx}

\newcommand{\myelips}[1]{\tikz \draw[fill=gray!60!white] (0,0) ellipse (28pt and 20pt);}

\begin{document}
%
\thispagestyle{empty}
\baselineskip = 18pt

\begin{center}

{\Large \bf  Probability Link Models with Symmetric Information Divergence}

 \vspace{.3in}

{\large \bf Majid Asadi$^a$, Karthik Devarajan$^b$,  Nader Ebrahimi$^c$,   Ehsan S. Soofi$^d$, \\
Lauren Spirko-Burns$^e$}

 \vspace{.2in}

{\small \em $^a$Department of Statistics,  University of Isfahan,    Iran,
and School of Mathematics, Institute for Research in Fundamental Sciences (IPM),  Tehran,  m.asadi@stat.ui.ac.ir

$^b$Department of Biostatistics and Bioinformatics, Fox Chase Cancer Center,
 Temple University Health System,  Philadelphia, PA,  karthik.devarajan@fccc.edu

$^c$Department  of Statistics, Northern Illinois University, DeKalb, IL,    nebrahimi@niu.edu

$^c$Sheldon B. Lubar School of Business,  University of Wisconsin-Milwaukee, Milwaukee, WI,\\   esoofi@uwm.edu

$^e$Department of Statistical Science, Temple University, Philadelphia, PA, LBurns@temple.edu}

 \today

\end{center}

\begin{abstract}
This paper introduces link functions for transforming one probability distribution to another such that the Kullback-Leibler and R\'enyi divergences between the two distributions are symmetric. Two general classes of  link models are proposed. The first model links two survival functions and is applicable to models such as the proportional odds and change point, which are used in survival analysis and reliability modeling. A prototype application involving the proportional odds model demonstrates advantages of symmetric divergence measures over asymmetric measures for assessing the efficacy of features  and for model averaging purposes. The advantages include providing unique ranks  for models and unique information weights for model averaging with one-half as much computation requirement of asymmetric divergences. The second model links two cumulative probability distribution functions. This model produces a generalized location model which are continuous counterparts of the binary probability models such as probit and  logit models. Examples include the generalized probit and logit models which have appeared in the survival analysis literature, and a generalized Laplace model and a generalized Student-$t$ model, which are survival time models corresponding to the respective binary probability models. Lastly, extensions to symmetric divergence between survival functions and conditions for copula dependence information are presented.
\end{abstract}


\section{Introduction}
Within the context of feature evaluation and selection, an information divergence measure between a candidate model that includes features and the null model provides a measure for evaluating the usefulness of features. Unlike asymmetric divergence measures, a symmetric divergence measure results in a unique set of ranks for candidate models. The set of divergences of candidate models from the null model provides weights for model averaging purposes according to the size of the divergence. A symmetric divergence measure provides a unique set of information weights which are data driven and, in turn, related to the Bayes factor (Kullback, 1959, p.5). The use of such divergence measures can be particularly useful when dealing with a large number of features and when selecting features for further study is of interest. It is also useful when different models with multiple features need to be compared with each other and evaluated. A practical advantage of symmetric divergence is that it reduces the computational burden by half, which can be substantial when selecting features for model building or when pairwise comparisons between models with subsets of features is under consideration.

This paper introduces broad families of probability models that provide symmetric  information divergences
 for evaluating  subsets of features. These link models are applicable
to continuous (Liu et al., 2019) and censored survival outcomes (Spirko-Burns \& Devarajan, 2020), examples of which are abundant
in the biomedical literature. These models also produce continuous versions of models used for binary data
(Mount et al., 2014; Nikooienejad et al., 2016) and ordinal data (Zhang et al., 2018).

Families of probability models are defined according to link functions of form $P_j(x)=G(P_k(x))$,
where $G(x)=P(X\leq x)$ is a cumulative distribution function (CDF) with a continuous probability density function (PDF) $g$
and $P$ represents either a survival function $S(x)=P(X>x)$ or a CDF $F(x)=P(X\leq x)$ with PDF $f$.
 The subscripts $k$ and $j$ stand for the dependency of the distributional parameters  to different subsets of features,
$\theta_j=\theta(V_j)$ and $\theta_k=\theta(V_k)$, where $V_j\subseteq \{z_1,\dots, z_p\}$.

The Kullback-Leibler (KL)  information divergence between two   PDFs is defined by
$$
K_{jk}=K(f_j:f_k) = \int f_j(x) \log \dfrac{f_j(x)}{f_k(x)}   dx   \geq 0,
$$
where the inequality becomes equality  if and only if $f_k(x)=f_j(x)$ almost everywhere,
provided that   $f_j$ is absolutely continuous with respect to $f_k$;
throughout this paper we assume that the two PDFs are absolutely continuous with respect to each other.

The KL divergence possesses some desirable properties and is used in various fields.
Despite its theoretical appeals, what becomes a nuisance in its applications  is the lack of symmetry in its arguments.
The lack of symmetry may be of no concern or even desirable in situations where a natural or an ideal reference is at hand.
But for applications to some important models, such as the  widely used proportional hazards model in survival analysis,
the order of the arguments in $K(f_j:f_k)$ can yield substantially different values.
For some other models, only one of the KL divergences between the two distributions involved may be available in closed form.
For example, it is shown
by Spirko (2017) that  one of the KL measures of the model proposed by Yang and Prentice (2005) has a closed form but the other
can be calculated only in terms of infinite series.

Historically, the lack of symmetry has been dealt with  by using Jeffreys divergence defined by
$$
J_{jk}=J(f_j,f_k) = K_{jk}+K_{kj}.
$$
Lin (1991) introduced the Jensen-Shannon divergence for the mixture of a set of $N$ distributions  defined as follows:
$$
JS(f_m;w)=\sum_{j=1}^N w_j K(f_j:f_m), \quad f_m(x)=\sum_{j=1}^N w_jf_j(x), \quad  w_j>0, \quad \sum_{j=1}^N w_j=1.
$$
This measure is symmetric for equal weights $w_j=1/N, j=1,\dots, N$.
Bernardo and Rueda (2002) defined the  intrinsic information as $\min \{K_{jk},K_{kj} \}$ for developing reference prior distributions.
Seghouane and Amari (2007) developed AIC-type criteria via symmetrizing the KL divergence by the arithmetic, geometric, and harmonic  means
of $K_{jk}$ and $K_{kj}$.

The desire for a symmetric information divergence between two PDFs, without a need to symmetrizing, has led to the use
 of R\'enyi information divergence defined by
$$
K_q(f_j:f_k) = \left \{\begin{array}{lll}
 \frac{1}{q-1}\log \int f_j^q(x)f_k^{1-q}(x)dx, & q >0, q\neq 1,\\ [.1in]
  K(f_j:f_k), & q=1,
\end{array} \right .
$$
where $K(f_j:f_k)=\lim_{q \uparrow 1}K_q(f_j,f_k)$.
In general, only  the case of $K_{1/2}$ is symmetric, which has been attractive for applications.
Hirschberg et al.\,(1991) and Granger et al.\,(2004) used  a normalized index defined by $1-\exp[-K_{1/2}(f:f_jf_k)]$, where $f$ is a bivariate PDF,
and $f_j, f_k$ are   marginal PDFs of $F$ for measuring dependence between two random variables associated with marginal PDFs.
These authors argued in favor of  $K_{1/2}(f:f_1f_2)$ vis \'a vis  the
mutual information between two random variables, $K(f:f_jf_k)$, which is widely used in many fields.

More recently, Nielsen (2019) extended the idea of symmetry of KL divergence in the normal family with a common variance,  $f_j=N(\mu_j, \sigma^2)$,  and gave a condition for symmetry of KL divergence for the location-scale family with a common scale.  Our proposed link functions provide broad families of probability models, including the location-scale family, where KL and R\'enyi divergences are symmetric.

In Section \ref{Sec:Link} we present two link functions.
Subsection \ref{Sec:PO} presents a survival link model, $S_j(x)=G(S_k(x))$, where  $G$ is a CDF with a PDF $g(u)>0$ on the unit interval.
We give  necessary and sufficient conditions for  symmetry of KL and R\'enyi divergences between the two corresponding PDFs in the link model.
Important cases of these survival link models are the proportional odds models (Marshall \& Olkin, 1997) and a class of change point models.
The CDF of logistic distribution is a PO model which was explicated by Doksum \& Gasko (1990) in terms of the survival model that corresponds
to  binary logistic regression.
PO models are useful for ordinal regression and classification problems with applications in preference learning, information retrieval, biomedicine,
weather prediction and economics (Perez-Ortiz et al., 2013; 2019).
Change-point models have applications in  many of the areas mentioned above involving longitudinal data; in particular,
for data generated by sensors such as in geology where data streams occur over depths. Eruhimov et al. (2007) outline
an approach that transforms the change-point problem into one of supervised feature selection.
It is also worth noting that survival models, and methods for survival and competing risks analysis in general,
have found non-standard applications in propagation and diffusion of contagions such as information, behaviors,
ideas and diseases as well as in information retrieval (Gomez-Rodriguez et al., 2013; Wu et al., 2018; Yin et al., 2013).

Subsection \ref{Sec:Location}  presents  the generalized location link model (GLL),
$ G^{-1}(F_j(x))=\theta + G^{-1}(F_k(x))$,
where $G$ is a CDF with a PDF $g(y)>0$ on the real line.
We give necessary and sufficient conditions for  symmetry of KL and R\'enyi  divergences between the two corresponding PDFs in the link model.
The GLL model with symmetric divergences includes the generalized probit model (Bagdonavicius \& Nikulin, 1997), which originally was explicated by
Doksum \& Gasko (1990) in terms of the survival model that corresponds to binary probit regression.
Probit regression is used in computer vision, cancer genomics,
natural language processing, social science, marketing, education and computational sustainability (Zheng \& Liu, 2012; Chen et al., 2018).
Other examples of GLL include logistic,  Laplace, and Student-$t$  CDFs which correspond to respective binary regression models.
These  models have symmetric PDFs; however, an example is constructed to show that the PDF symmetry  is sufficient but not necessary for the symmetry of divergence measures.
Location models such as the generalized Laplace and Student-t can be useful for anomaly detection due to their heavier tails.
The binary regression of these models combined with methods for sparse, deep and multi-task learning, have been found particularly useful
in applications where interactions of multiple entities or multi-task learning is of interest (Chen et al., 2018; Chang et al., 2019; Vargas et al., 2019; Wang \& Zhu, 2019).
The GLL provides models for applications where the distribution of a continuous random variable is of interest rather than a binary variable.

Section \ref{Sec:Cov} illustrates important consequences of using symmetric divergence for
data analysis using a prototype application involving the PO model.
In particular, it presents model evaluation and information model averaging that have distinct advantages
when a symmetric divergence is utilized.
This section also  gives a comparison between the symmetric divergence of the PO model
with the null model and the asymmetric divergences of the more widely used Cox proportional hazards model in terms of feature effect.

Section \ref{Sec:Extensions} discusses two extensions pertaining to the transformations of the survival function.
The first extension pertains to    survival function $S_j$ scaled by its mean $\mu_j=\int_0^\infty S_j(x) dx<\infty$. This gives
$f^e_j(x)  = S_j(x)/\mu_j, x\geq 0$, which is  the PDF of the equilibrium distribution (ED) of $S_j$.
  The random variable associated with the ED, denoted as $A_X$, is referred to as the  asymptotic age (Shaked and Shanthikumar 2007)
due to the fact that  $E_j(A_X)$ is the expected asymptotic age of the renewal process at an age $t$ (see Ross 1983).
 The ED plays an important role in renewal processes.
Our result for survival transformation is extended to the equilibrium distribution, where $S^e_j(x)=G(S^e_k(x))$.
This extension provides symmetric  divergence measures in terms of the two scaled survival functions, $S_j$ and $S_k$.
The second extension pertains to the symmetry of the dependence information divergence. A result gives
the necessary and sufficient condition for symmetry of the dependence information divergence in terms of copula.

\section{Link models with symmetric divergence} \label{Sec:Link}
We refer to a CDF $G$ with a continuous PDF $g$ as a link function when it is used to transform a baseline survival function or a CDF to another survival function or CDF.
This section provides conditions for the  symmetry of  the KL and R\'enyi divergences between the PDF of the baseline model $f_1$ and
the PDF of the transformed model $f_2$.

\subsection{Survival link models} \label{Sec:PO}
The following proposition gives a necessary and sufficient condition for the symmetry of the KL divergences for a  transformation of the survival function.

\begin{Proposition} \label{Pro:K12=K21S}
Let $S_1(x)=G(S_2(x))$, where $G$ is a  continuous  CDF with PDF $g(u)$ on the unit interval.
 Then $K_q(f_1,f_2) =K_q(f_2,f_1)$ if and only if $K_q(g:g^*) =K_q(g^*:g)$ for all $q>0$, where $g^*$ is the uniform PDF on $[0,1]$.
\end{Proposition}

The Shannon entropy of a continuous distribution with PDF $g$ is defined by
$$
H(g)=-\int g(u)\log g(u) du,
$$
provided that the integral converges.
Noting that $K (g:g^*)=-H(g)$ and $K (g^*:g)=-E^*[\log g(U)]$, where $E^*$ denotes the expectation with respect to the uniform PDF $g^*$,
we have the following equivalent condition in terms of the entropy: $K (f_1:f_2)=K (f_2:f_1)$ if and only if  $H(g)= E^*[\log g(U)]$.
However, for the R\'enyi divergence, the entropy representation is only defined for $q\leq 1$ (see Appendix).

A family of survival transformation models studied by Marshall \& Olkin (1997) is
\begin{equation}
\label{Eq:Gu}
G(x)=\frac{\alpha S_2(x)}{1-\bar \alpha S_2(x)},  \quad  \alpha >0; \ -\infty < x < \infty ,
\end{equation}
where $\bar \alpha=1-\alpha$ and $\alpha>0$ is free from $x$ and called   the tilt parameter.
The baseline and transformed distributions are proportional odds (PO), defined by
$\psi_1(x)=\alpha \psi_2(x)$, where $\psi_i(x)=\frac{F_i(x)}{S_i(x)}, i=1,2$.

\begin{Corollary} \label{Cor:PO}
For the PO family (\ref{Eq:Gu}), $S_1(x)=G(S_2(x))$ where the CDF $G$ is the PO model (\ref{Eq:Gu}),
Then the condition of Proposition  \ref{Pro:K12=K21S} is satisfied and
$$
K_q(g:g^*) = K_q(g^*:g) =\frac{1}{q-1} \log \frac{\alpha^q-\alpha^{1-q}}{(\alpha-1)(2q-1)},  \quad  q>0, \ q\neq 1/2.
$$
The  log term  converges as $q\to1/2$.
\end{Corollary}

The following example gives a   application of Corollary \ref{Cor:PO}.

\begin{Example}   \rm
Let $X_1,\dots,X_N$  be independent and identically distributed with survival function $S_2$ where $N$ is a random variable independent of $X_i$'s
with the geometric distribution
$$
p_n=p(1-p)^{n-1},  \quad  n=1,2,\dots; \ 0<p<1 .
$$
Then the distributions of extrema $X_{1:N}=\min(X_1,\dots,X_N)$ and $X_{N:N}=\max(X_1,\dots,X_N)$ are  in the PO family $G$
in (\ref{Eq:Gu}) with parameters $\alpha=p$  and $\alpha=1/p$, respectively (Marshall \& Olkin, 1997).
Thus, $K_q(f,f_{\min})=K_q(f_{\min},f)$ and $K_q(f,f_{\max})=K_q(f_{\max},f)$ for all $q>0$, where $f_{\min}$ and $f_{\max}$
are PDFs of $X_{1:N}$ and $X_{N:N}$, respectively. An example of geometric extreme stable distribution is the logistic distribution.
If $S_2(x)=(1+e^{\lambda x})^{-1}, x\in \Re ,  \lambda >0$, then the PO link gives
$$
S_1(x)= \frac{1}{1+e^{\eta+\lambda x}}, \quad  x\in \Re , \ \eta=-\log\alpha, \ \alpha,\lambda >0.
$$
\end{Example}

Applications of Proposition \ref{Pro:K12=K21S} are not limited to the PO models.
The following example illustrates this fact.

\begin{Example}   \label{Ex:PW}  \rm
Let $u=S(x)$ and $G$ be the CDF of the following piecewise uniform family:
\begin{equation}
\label{Eq:GPW}
G(u) =\frac{(1-p)u}{p} \delta ( 0 \leq u \leq p) + \left [1-p+\frac{p(u-p)}{1-p}\right ] \delta( p \leq u \leq 1), \quad  0\leq u\leq 1; \ 0<p<1 ,
 \end{equation}
where $\delta(A)$ is the indicator function of set $A$.
Then for any baseline survival function  the PDF of $S_2(x)=G(S_1(x))$ satisfies the condition of Proposition  \ref{Pro:K12=K21S}.
 Clearly, $G$ is not in the PO family but satisfies Proposition  \ref{Pro:K12=K21S}.
Like the PO model,   link function (\ref{Eq:GPW}) generates piecewise parametric  distributions.
For example consider the exponential model $S_2(x)=e^{-\lambda x}, x \geq 0$. Using this in (\ref{Eq:GPW}) gives the following piecewise exponential survival function:
\begin{eqnarray*}
S_1(x) =\left [1-p + \frac{p}{1-p} \left (e^{-\lambda x}-p\right )\right ] \delta \left (0 \leq  x \leq  x_p \right )
+ \frac{1-p}{p} e^{-\lambda x} \delta \left (  x \geq x_p \right ),  \quad   0<p<1 ,
  \end{eqnarray*}
where $x_p=-\lambda^{-1}\log p$ is the change point.
 \end{Example}

\subsection{Generalized location link models} \label{Sec:Location}
The location-scale family is defined by PDFs, $f_i(x)= \sigma_i^{-1}f ((x-\mu_i)/\sigma_i), i=1,2$,
where $f$ is in the same family as $f_i$ with location zero and scale one.
The entropy is location-invariant implying that in a location-scale family, $K (f_1:f_2)=K (f_2:f_1)$ if and only if
$$
  E_1[\log f_2(X)] -E_2[\log f_1(X)]=\log \frac{\sigma_1}{\sigma_2},
$$
where $E_i$ denotes the expectation with respect to $f_i$.
 A condition given by Nielsen (2019) for the symmetry of the KL divergence for the location-scale family with a common scale $\sigma$ can be represented as follows:
\begin{equation}
\label{Eq:E1E2c}
  E_f[\log f(X+m)] =E_f[\log f(X-m)],
\end{equation}
where $m=(\mu_1-\mu_2)/\sigma$.
Clearly, this condition is satisfied by symmetric location families with location $c$ and a common scale.
However, this condition may not hold for an asymmetric location family with a common scale.
The following example illustrates  this fact.

\begin{Example} \rm \
Consider the family of Gumbel (generalized extreme value distribution Type-I) distributions with PDFs
$$
f_\mu(x)=e^{-\left [(x-\mu)+e^{-(x-\mu)}\right]} .
$$
 Using $E_f(X)=\mu+\gamma$
and the moment generating function of the family, $E_f(e^{tX})=\Gamma(1-t)e^{\mu t}$, we find that
(\ref{Eq:E1E2c}) holds for the root of $e^m-e^{-m}-2m =0$. However, the only solution of this equation is $m= 0$.

\end{Example}

The generalized location   random variables associated by $F_i, i=1,2$, is defined by
\begin{equation}
\label{Eq:GinvLL}
 G^{-1}[F_1(x)] =  G^{-1}[F_2(x)]+\theta,   \quad  -\infty < x < \infty ,
\end{equation}
where $G$ is a continuous CDF with support $\Re$.
Well-known examples of generalized location models are the generalized probit and generalized logit.

 Let $Y_i= G^{-1}[F_i(X)], i=1,2$, and $Y_1 =_{st} Y_2+\theta$, where ``$=_{st}$" denotes the stochastic equality.
Denote the distribution of $Y_1$ and $Y_2+\theta$ by $G_1$  and $G_2$, respectively. These distributions are related as
$G_1(y) = G_2(y+\theta)$ and define  a location family   with  location parameters  related by $\theta_2=\theta_1+\theta$.
In the generalized location model (\ref{Eq:GinvLL}) the CDF $G$ transforms $F_2(x)$ to $F_2(x)$ as follows:
\begin{equation}
\label{Eq:GLL}
F_1(x) = G[G^{-1}(F_2(x))+\theta],  \quad  -\infty < x < \infty .
\end{equation}
The following proposition gives  conditions for KL and R\'enyi divergence symmetry of (\ref{Eq:GLL}).

\begin{Proposition} \label{Pro:K12=K21F}
Let $g$ be the PDF of the link $G$ in (\ref{Eq:GLL}). Then:
\begin{enumerate}
  \item[(a)]
$K(f_1:f_2)=K(f_2:f_1)$ if and only if $E_g[\log g(X+\theta)]=E_g[\log g(X-\theta)]$, where $E_g$ denotes the expectation with respect to $g$.
  \item[(b)]
$K_q(f_1:f_2)=K_q(f_2:f_1)$ if and only if $K_q(g(x):g(x+\theta))=K_q(g(x):g(x-\theta))$ for all $q>0$.
\end{enumerate}
\end{Proposition}

Clearly, the PDF of the symmetric location families satisfies the conditions of Proposition \ref{Pro:K12=K21F}.
The following example illustrates applications of Proposition \ref{Pro:K12=K21F} in survival analysis.

\begin{Example} \rm \
The generalized location model relates the effects of a vector of features $z$ on a probability distribution of the time $t$ to an event
additively as follows:
\begin{equation}
\label{Eq:PGProbit}
G^{-1}(F(t\mid z))= G^{-1}(F_0(t))+\beta^\prime z,
\end{equation}
\begin{enumerate}
  \item[(a)]
The generalized probit model is given by the standard normal CDF $G(y)=\Phi(y)$ in (\ref{Eq:PGProbit}).
  \item[(b)]
The  generalized logit model is given by the logistic CDF $G(y)=(1+e^{-y})^{-1}$ in (\ref{Eq:PGProbit}).
    \item[(c)]
The Student-$t$ and Laplace distributions are scale mixtures of normal distribution and enable  capturing observations far from the center.
The generalized Student-$t$ and generalized Laplace location models are given by using the respective CDFs for $G$ in (\ref{Eq:PGProbit}).
\end{enumerate}
\end{Example}

The symmetry of the location family is not necessary for the conditions of Proposition \ref{Pro:K12=K21F}.
The following example illustrates this fact.

\begin{figure}
  \centering
    \includegraphics[scale=.4]{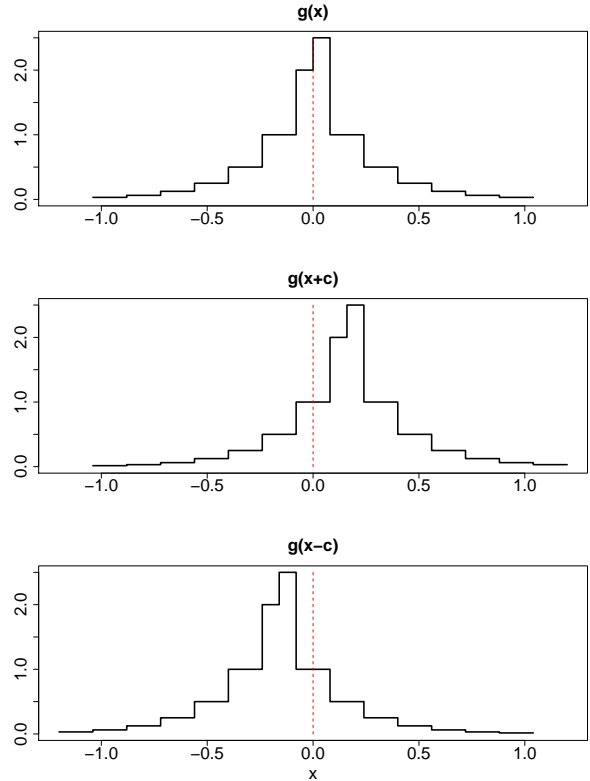}
  \caption{Plots of asymmetric piecewise uniform PDF $g(x)$ for $\theta=0.16,  a_1=2, a_2=2.5$, and $b_k=2^{-(k-1)}$. }
  \label{fig:Asymm-L}
\end{figure}

\begin{Example} \label{Ex:Asymm} \rm
Let
$$
g(x)=a_1\delta(-\theta/2 \leq x < 0))+a_2 \delta(0\leq x < \theta/2)+b_k \delta(k \theta/2 < |x| < (2k+1) \theta/2),
$$
where $a_i>0, i=1,2$ and $0< b_k \leq 1, k=1,2, \dots$ are such that
$$
 0 < \frac{1}{\theta}- \frac{a_1+a_2}{2}=  2 \sum_{k=1}^\infty b_k <\infty .
$$
The distribution is  asymmetric when $a_1\neq a_2$, but $g(x)$ satisfies  the condition of Proposition \ref{Pro:K12=K21F}.
This is seen by noting that for $-\theta/2 < x < \theta/2, g(x+\theta)=g(x-\theta)$ and for the symmetric parts of $g(x), g(x+\theta)-g(x-\theta)=g(x-\theta)-g(x+\theta)$.
For example, let $b_k=n^{-(k-1)}, n>1$. Then,  $\theta\in (0,1)$ and  $ a_i>0, i=1,2$ can be chosen such that
$$
\frac{a_1+a_2}{2} = \frac{1}{\theta}-\frac{2n}{n-1} .
$$
Figure \ref{fig:Asymm-L} shows plots of $g(x)$  for  $n=2, \theta=0.16, a_1=2$ and $a_2=2.5$.
\end{Example}

\section{Model evaluation and averaging} \label{Sec:Cov}
In preceding sections we provided several application examples for the survival transformation and generalized location models.
In this section, we illustrate applications to feature evaluation.

In applications  the tilt parameter of the PO model $\alpha$ and the location parameter of the generalized location model  $\theta$
are modeled in terms of a set of features $\{z_1,\dots, z_p\}$. The feature selection for prediction involves evaluating information
values of subsets of features. For the PO model  $\alpha=\exp(\beta^\prime z)$ and for the generalized location model  $\theta=\beta^\prime z$.
Let $V_j\subseteq \{z_1,\dots, z_p\}, j=0, 1,\dots, N \leq 2^p-1$, where $V_0=\emptyset$,
and $\beta_j$ denote the vector of respective feature coefficients. Each $V_j$ provides a model $M_j$.
Pairs of models can be compared by
$$
K_{jk}=K_q(M_j:M_k), \quad  k\neq j, \ j=1,\dots,N.
$$
The divergence with the null model $M_0$, $K_{j0}$, provides a measure for evaluating usefulness of features for prediction.
A symmetric divergence, where $K_{jk}=K_{kj}$, has important consequences for applications.
\begin{enumerate}
  \item[(a)]
A symmetric divergence requires $N/2$ pairwise comparisons.
  \item[(b)]
Divergence measures can be used for ranking models. With an asymmetric divergence, the ranks of $K_{jk}$ can be different from
the ranks of $K_{kj}$. A symmetric divergence provides a unique set of ranks.
  \item[(c)]
The set of divergences with the null provides the following weights for model averaging purposes:
\begin{equation}
\label{Eq:IW}
w_j=\frac{K_{i0}}{\sum_{j=1}^J K_{j0}}.
\end{equation}
This is the information weighting scheme, in that subsets are given weights according to the expected information they provide against the null;
i.e., the expected logarithm of Bayes factor (Kullback, 1959, p.5).
The symmetry of the KL divergence for models implies that  the expected logarithm of Bayes factor can also be symmetric.
With an asymmetric divergence, the weights given by $K_{jk}$ can be different from
the weights given by $K_{kj}$. A symmetric divergence provides a unique set of weights, hence unique consequent model averaging results
by the information weighting scheme.
\end{enumerate}

The  information weights (\ref{Eq:IW}) provide the following average predictive model
\begin{equation}
\label{Eq:MW}
S_m(x)= \sum_{j=1}^J w_i S_{j0}(x).
\end{equation}
Symmetry of divergence is not closed under mixing. Also divergence measures involving a mixture are not readily available.
The informativeness of the mixture model relative to a reference model with PDF $f_r$ can be assessed by the following inequality:
\begin{equation}
\label{Eq:Kmr}
K(f_m:f_r)\leq \sum_{j=1}^N w_j K_{jr};
\end{equation}
this inequality is given in Kullback (1959, p. 23). For example, $f_r$ can be the null model or a model whose parameter is
given by the  average of regression parameters:
\begin{equation}
\label{Eq:Beta-r}
\beta_r=\sum_{j=1}^J w_j \beta_j .
\end{equation}

The Jensen Shannon divergence of the mixture gives the average of divergences between all pairs of constituents of the mixture model.
This measure is bound as follows:
\begin{equation}
\label{Eq:JS}
JS(f_m; w)= \sum_{j=1}^N w_j K_{jm} \leq \min \left \{B(w,J_{kj}), H(w)  \right \},
\end{equation}
where $H(w)=-\sum_{j=1}^J w_j \log w_j$ is the entropy of the weight distribution,
$$
B(w,J_{kj})=\sum_{j<k}^N w_j w_k J_{jk},
$$
and $J_{jk}$ is the Jeffreys divergence;
the inequality in terms of $H(w)$ is shown in Wang and Madiman (2014) and in terms of $B(w,J_{kj})$
is given in Kullback (1959, p. 23) and is shown in Asadi et al.\,(2016).
The symmetry of divergence $K_{jk}$ gives $J_{jk}=2K_{jk}$ and provides 50\% reduction for computing $B(w,J_{kj})$.

\subsection{PO versus PH models}
This section compares the PO model with  a model where the divergence is not symmetric.
The hazard rate corresponding to a survival function $S_i$ is defined by  $r_i(x)=f_i(x)/S_i(x)$.
The proportional hazards (PH) model is defined by $r_1(x)=\pi r_2(x), \pi>0$ and represented as $S_1(x)=S_2^\pi(x)$.
Unlike the PO model,  the PH family   does not satisfy Proposition  \ref{Pro:K12=K21S} for $q\neq 1/2$.
The PO model is an alternative to the more widely used  PH  model in survival analysis
where the effects of features $z$ is also captured by   $\pi=\exp(\beta^\prime z)$ (Cox, 1972).
But unlike the PO model,  the PH family   does not satisfy Proposition  \ref{Pro:K12=K21S} for $q\neq 1/2$.
For PH models
\begin{eqnarray*}
&&  K(f_\pi:f_0)= e^{-\beta^\prime z} +\beta^\prime z -1,   \quad \pi\neq 1,\\
&&  K(f_0:f_\pi)= e^{\beta^\prime z} -\beta^\prime z -1, \quad \pi\neq 1,
\end{eqnarray*}
where $f_0(x)=f_{\pi=1}(x)=f_{\beta=0}(x), x \in \Re^p$.

\begin{figure}
  \centering
   \includegraphics[scale=.8]{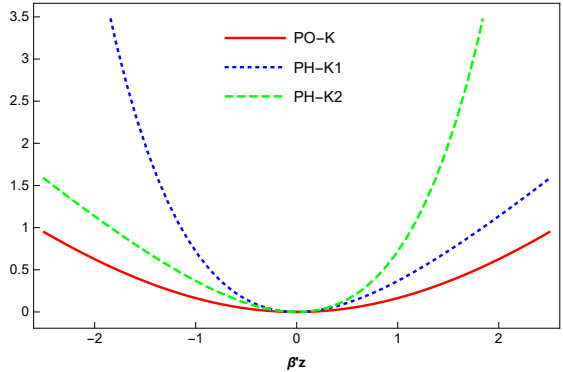}
 \caption{Effects of features $\beta^\prime z$ on the KL divergence between two proportional odds models (PO-K) and between two proportional hazards models (PH-K1=$K(f_\pi :f)$, PH-K2=$K(f:f_\pi))$.}
  \label{fig:PO-K}
\end{figure}

Figure \ref{fig:PO-K} compares the  KL divergence $K(f_\alpha, f_0)$ for the PO model with tilt parameter $\alpha=\exp(\beta^\prime z)$,
where $f_0(x)=f_{\alpha=1}(x)=f_{\beta=0}(x), x \in \Re^p$ and the two KL divergences for the PH models.
These plots are visualizations in terms of $\beta^\prime z$ and illustrate  some interesting features of  these measures.
The plot for the PO model shows that the KL is symmetric about $\beta^\prime z=0$, which indicates no effect of the feature.
The plots for the PH model show  symmetry between the two KL divergences: $(f_\alpha:f_1)$ for $\beta^\prime z< (>) \ 0$ is equal
to the $K(f_1:f_\pi)$ for $\beta^\prime z > (<) \ 0$.
The plots  show that the two KL divergences for the PH model are substantially different when the effects of features are strongly positive or negative. Also, the KL measure for the PO model is dominated by both KL measures for the PH model.

\subsection{A prototype application}
Data  is from the Mayo Clinic trial in primary biliary cirrhosis (PBC) of the liver conducted between 1974 and 1984.
It is a publicly available data set in R, which contains 418 observations in its original form.
It contains information on survival time (which is the observed or censored survival time) and status
(which indicates if the observation has been censored).
Following Martinussen and Scheike (2006) we use the following five features and log-transformations for albumin, bili and protime
Age ($z_1$), Edema ($z_2$), logAlbumin ($z_3$), logBili ($z_4$) and logProtime ($z_5$).
 Subjects that received transplants were removed and 393 cases are used in our analysis.
We evaluate PO models based on all combinations of features at the median point continuous features for patients with and without presence of Edema:
$$
z^\prime = (51.92, 0, 1.26, 0.26, 2.36), \quad z^\prime = (51.92, 1, 1.26, 0.26, 2.36).
$$

The PO model (\ref{Eq:Gu}) assumes that the effect of features   proportionately increases or decreases the odds of failure of an item or recurrence at time $z$.
  Consider the PO models $S_1=G(S_0(x))$ with a baseline model $S_0$ and tilt parameter $\alpha=\exp(\beta^\prime z)$.
It is shown by Spirko-Burns and Devarajan (2020) that the KL divergence between the PO models $f_\alpha=f_{\alpha \neq 1}$ and the null model  is symmetric:
\begin{equation}
\label{Eq:Kj0}
K(f_\alpha:f_0)= K(f_0:f_\alpha)= \frac{e^{\beta^\prime z}+1}{e^{\beta^\prime z}-1} \beta^\prime z-2, \quad \alpha \neq 1,
\end{equation}
where $f_0(x)=f_{\alpha=1}(x)=f_{\beta=0}(x), x \in \Re^p$. Corollary \ref{Cor:PO} extends this result to the R\'enyi divergence.

Let $M_j$ be the PO model $S_{j0}(x)=G_j(S_0(x))$, where $S_0=S_{\alpha=1}=S_{\beta=0}$.
Then $K_{j0}$  is given by the KL between the PDFs of  $S_{j0}$ and $S_0$.
We obtain the KL divergence between a PO model  $M_j$, with respect to a reference  PO model $M_r$ with $S_{r0}(x)=G_r(S_0(x))$.
Taking the reference PO model for the base we have, $S_{jr}(x) =G_j(S_{r0}(x))$ and the tilt parameter $\alpha_{jr}$ is found in terms of
$\alpha_{j0}$ and $\alpha_{r0}$ as follows:
 \begin{eqnarray*}
&&\frac{F_{r0}(x)}{S_{r0}(x)} \equiv \frac{F(x|\alpha_{r0})}{S(x|\alpha_{r0})}= \frac{1}{\alpha_{r0}}\frac{F_0(x)}{S_0(x)}, \\
&&\frac{F_{jr}(x)}{S_{jr}(x)} \equiv \frac{1}{\alpha_{jr}}\frac{F_{r0}(x)}{S_{r0}(x)}
=\frac{1}{\alpha_{jr}\alpha_{r0}}\frac{F_0(x)}{S_0(x)}=\frac{1}{\alpha_{j0}}\frac{F_0(x)}{S_0(x)} .
\end{eqnarray*}
This gives $\alpha_{jr}=\alpha_{j0}/\alpha_{r0}$.
The  KL between the PDFs of   two PO models $S_{ir}$ and $S_{r0}$ is
\begin{eqnarray}
\label{Eq:Kjr}
K_{jr}= K(f_{jr}:f_{r0})
&=&\frac{\alpha_{j0}+\alpha_{r0}}{\alpha_{j0}-\alpha_{r0}}\log \frac{\alpha_{j0}}{\alpha_{r0}} -2\cr
 &=&\frac{e^{\beta_j^\prime z}+e^{\beta_r^\prime z}}{e^{\beta_j^\prime z}-e^{\beta_r^\prime z}}(\beta_j^\prime z- \beta_r^\prime z)-2 .
\end{eqnarray}

\begin{table}
\caption{\label{TAB:PBC1} Information analysis of all subsets of five features of PBC data.}
$$ \small
\begin{array}{clrrrrrrr}
&& \multicolumn{3}{c}{\mbox{Edema absent, }z_2=0} &  & \multicolumn{3}{c}{\mbox{Edema present, } z_2=1} \\
 j&  \mbox{Covariates}  & \hat \beta_j^\prime z  & K_{j0} & \mbox{Rank}    &  & \hat \beta_j^\prime z  & K_{j0}  & \mbox{Rank}  \\ [.05in]
\hline
1& z_1 &2.305& 0.816    &25&&&&28\\
2& z_2 &0.000& 0.000    &31&&3.170&1.448&24\\
3& z_3 &-8.753& 6.756    &14&&&&15\\
4& z_4 &0.354& 0.021    &29&&&&31\\
5& z_5 &22.263& 20.263   &2&&&&2\\ [.1in]
6& z_1, z_2 &1.739& 0.480    &27&&4.679&2.767&21\\
7& z_1, z_3 &-6.464& 4.484    &18&&&&18\\
8&z_1, z_4 &3.890& 2.053    &23&&&&23\\
9& z_1, z_5 &23.896& 21.896   &1&&&&1\\
10& z_2, z_3 &-6.824& 4.838    &16&&&&22\\
11& z_2, z_4 &0.323& 0.017    &30&&2.660&1.060&26\\
12& z_2, z_5 &16.904& 14.904   &4&&19.388&17.388&4\\
13& z_3, z_4 &-5.913& 3.945    &19&&&&19\\
14& z_3, z_5 &12.128& 10.128   &8&&&&11\\
15& z_4, z_5 &13.701& 11.701   &7&&&&8\\ [.1in]
16& z_1, z_2, z_3 &-5.100& 3.162    &21&&-2.903&1.240&25\\
17&z_1, z_2, z_4 &3.288& 1.543    &24&&5.224&3.281&20\\
18& z_1, z_2, z_5 &18.358& 16.358   &3&&20.642&18.642&3\\
19&z_1, z_3, z_4 &-1.875& 0.554    &26&&&&29\\
20& z_1, z_3, z_5 &13.734& 11.734   &6&&&&6\\
21& z_1, z_4, z_5 &15.240& 13.240   &5&&&&5\\
22& z_2, z_3, z_4 &-4.416& 2.524    &22&&-2.593&1.013&27\\
23&z_2, z_3, z_5 &9.521& 7.522    &12&&11.301&9.302&12\\
24&z_2, z_4, z_5  &10.165& 8.166    &11&&12.289&10.289&10\\
25& z_3, z_4, z_5 &6.764& 4.780    &17&&&&17\\ [.1in]
26& z_1, z_2, z_3, z_4 &-1.073& 0.188    &28&&0.448&0.033&30\\
27&z_1, z_2, z_3, z_5  &10.981& 8.981    &10&&12.611&10.611&9\\
28&z_1, z_2, z_4, z_5 &11.961& 9.961    &9&&13.724&11.724&7\\
29&z_1, z_3, z_4, z_5 &9.067& 7.069    &13&&&&13\\
30&z_2, z_3, z_4, z_5 &5.428& 3.476    &20&&7.047&5.059&16\\ [.1in]
 31  &  z_1, z_2, z_3, z_4, z_5  &7.672& 5.679  &15&&9.040&7.042&14\\  [.1in]
 f_r & \mbox{Average }  \beta_r &12.286 &10.286 &&& 13.376&11.376&\\
 K_{mr} &  \mbox{Bound} & & 11.977  &&&   &12.005  \\  [.05in]
\hline
\end{array}
$$
\end{table}

Table \ref{TAB:PBC1} presents the information analysis for all subsets $V_j, j=1,\dots,31$.
For each prediction point, the table gives $\hat \beta_j^\prime z$ and the divergences between the implied PO model and null model, $K_{j0}$.
Ranks  for the models  are also shown in the table.
The last two rows of the table pertain to the PO model $f_r$ with $\alpha_r$ given by the average regression parameter (\ref{Eq:Beta-r})
by the information weights (\ref{Eq:IW}).
In this case, the tilt parameter of $S_{r0}$ is the geometric mean  $\alpha_{r0}=\prod_{j=1}^J\alpha_{j0}$.
The table gives the divergence  between $f_r$ and the null model when Edema is absent and present.
The last row  gives the average $K_{j0}$. This measure is the bound (\ref{Eq:Kmr}) for divergence between the average PO model $f_m$ (mixture) and $f_r$.
In both cases the bound is substantial.

The following points are noteworthy.
\begin{enumerate}
  \item[(a)]
For both prediction points the most informative single feature model is $V_5=\{z_5\}$ followed
by $V_3=\{z_3\}$ with a huge information difference, which is followed by  $V_1=\{z_1\}$ with a substantial lower information.
However,  $V_9=\{z_1, z_3\}$, is the most informative subset. The most informative subsets of size three and four contain $V_9$.
About half of the subsets are more informative than  full model that contains all features.
\item[(b)]
For both prediction points the PO models with parameters estimated by the average regression are more informative than the full model.
For the prediction point in the case when Edema is not present, the PO model with parameters  estimated by the average regression is
also more informative than the models with four features.  For the case when Edema is present,
the PO model with parameters  estimated by the average regression is almost equally informative
with most informative  model with four features.
\end{enumerate}

\begin{figure}
  \centering
  \includegraphics[scale=.4]{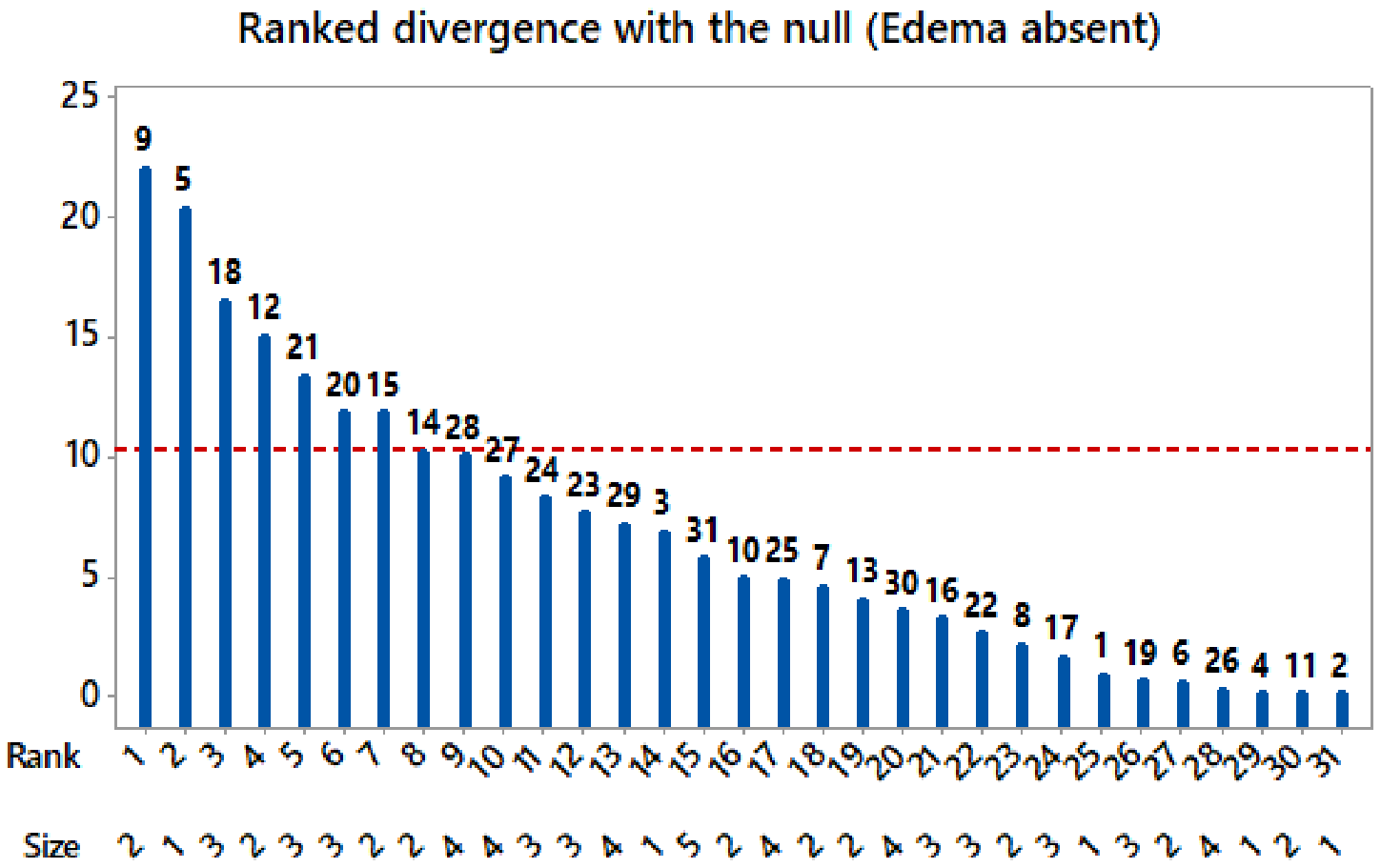}
  \includegraphics[scale=.4]{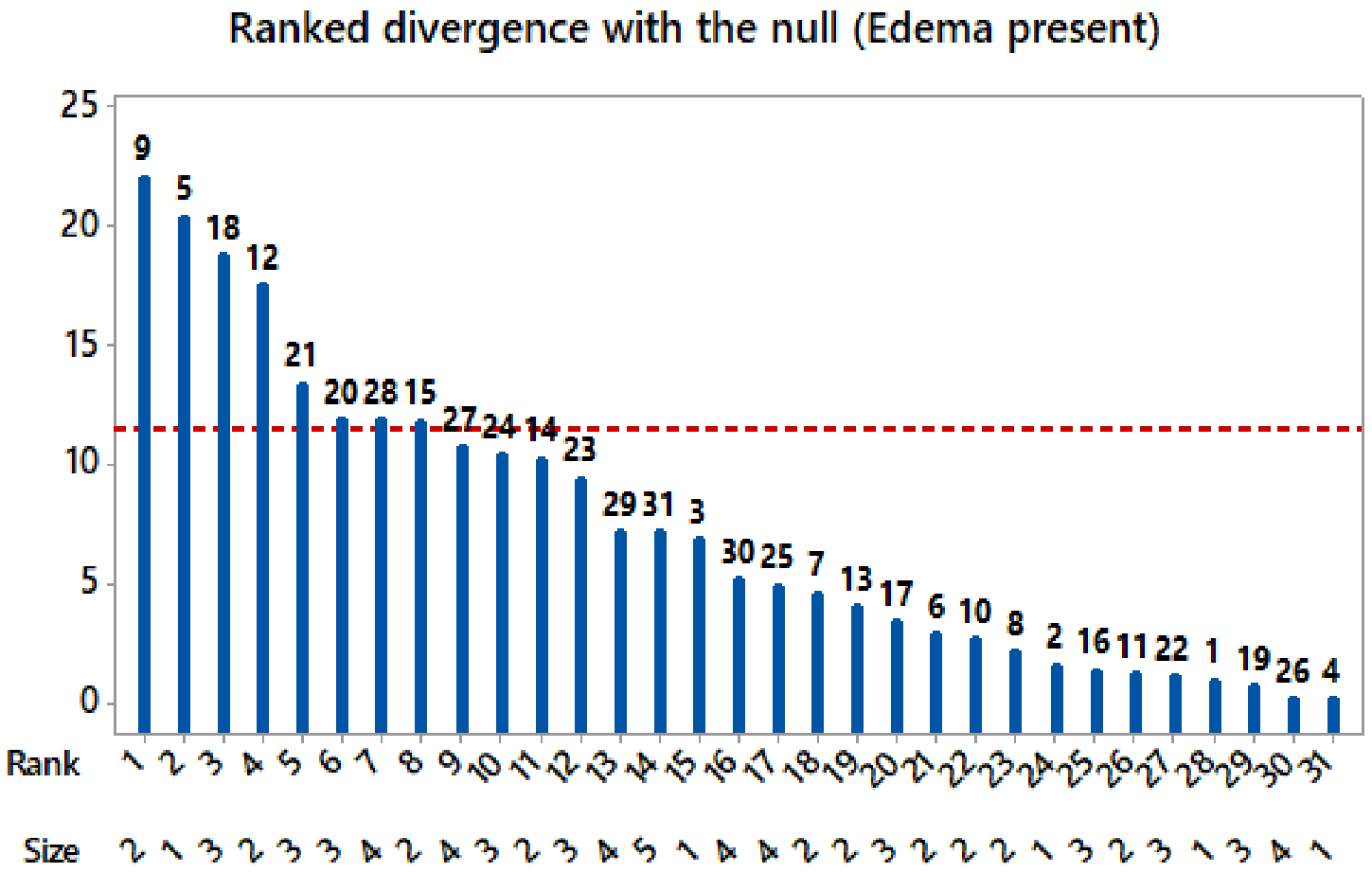}

\begin{picture}(50,30)
 \setlength{\unitlength}{.05in}
\scriptsize
   \put(-14,26){\textcolor{red}{$\bm{K_{r0}=10.286}$}}
   \put(35,26){\textcolor{red}{$\bm{K_{r0}=11.376}$}}
\end{picture}

\vspace{-.5in}
  \caption{Ranked information divergences of PO models with the null model for two predictive points; bar label identifies $M_j$ and size is the number of predictors in the model; dotted lines show the divergence between PO model with the average regression parameters and the null model.}
  \label{fig:PBC-PO-Kj0}
\end{figure}

Figure \ref{fig:PBC-PO-Kj0} gives a visualization of the information measures of  Table \ref{TAB:PBC1}.
The information measures for each predictive point are ranked from the highest to the lowest, depicted by the height of the bar.
The bar label identifies the model $M_j$ and the number of predictors in the model is shown along with the rank on the horizontal axis.
The dotted line shows the divergence between the PO model with the average regression parameter $\beta_r$ and the null model.

\begin{figure}
  \centering
    \includegraphics[scale=.4]{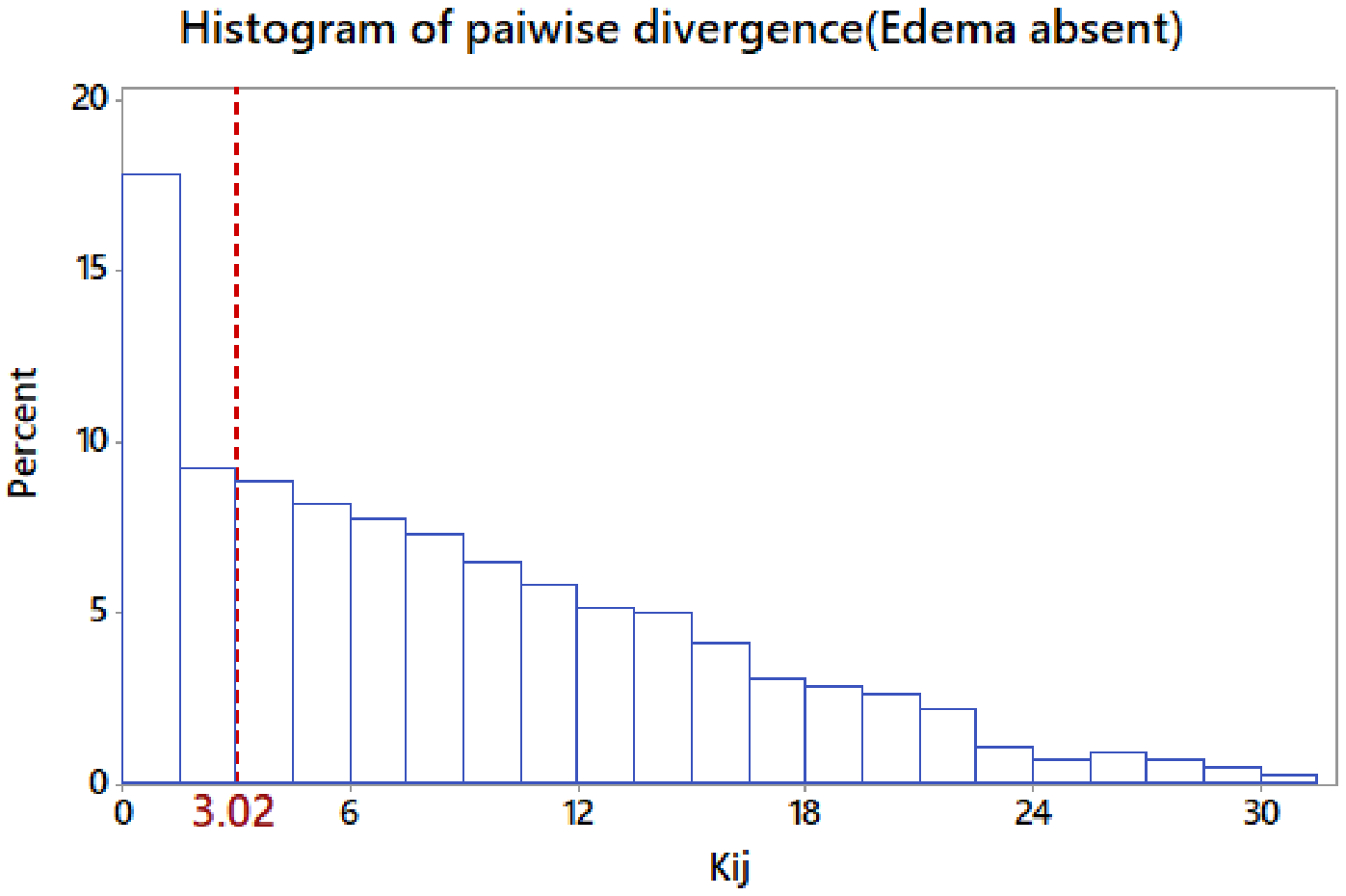}
    \includegraphics[scale=.4]{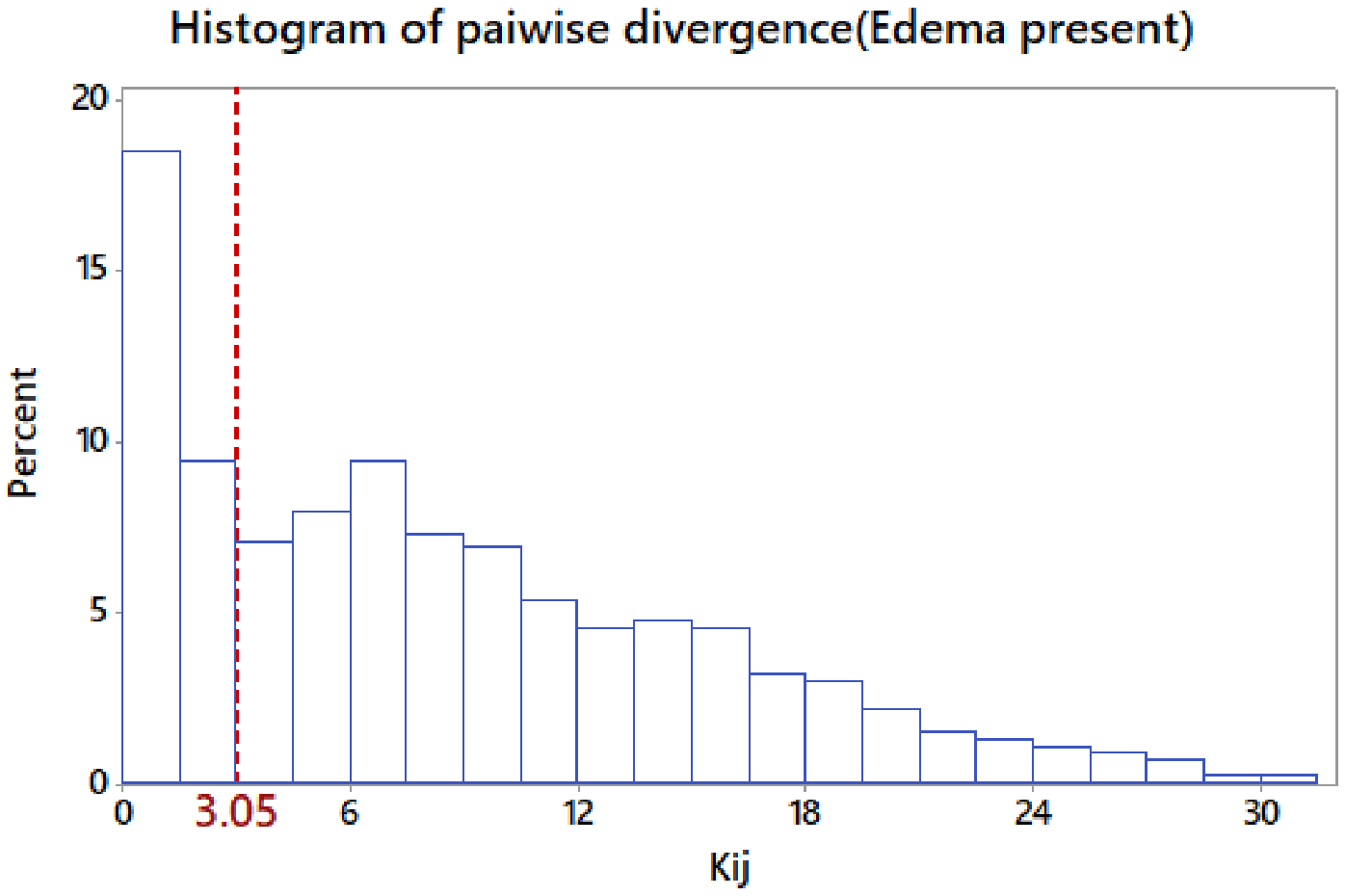}

\vspace{-.1in}

  \caption{Histograms of pairwise divergences $K_{ij}$ of all pairs of PO models for PBC data.}
  \label{fig:PBC-PO-Kij}
\end{figure}

Figure \ref{fig:PBC-PO-Kij} shows histograms of divergences between all pairs of PO models at prediction points where Edema is absent and present.
Each panel represents the distribution of $N(N-1)/2=365$ pairwise divergences. The dashed vertical lines are the minimum bound in (\ref{Eq:JS}) for the average of
$K_{jk}$'s given by $H(w)$ for both cases;  ($B(w,J_{ij})=7.985,  7.305$ for divergences where Edema is absent and present).
In each case about 28\% of the pairwise divergences are less than the bound.

\section{Extensions of survival transformation} \label{Sec:Extensions}
We provide two extensions of Corollary (\ref{Cor:PO}),

\subsection{Equilibrium survival model} \label{Sec:KLs}
Consider the following survival transformation:
\begin{equation}
\label{Eq:GSED}
S^e_1(x)=G(S^e_2(x))
\end{equation}
where $S^e_i$ is the survival function associated with $f^e_i$ and $G$ is a  continuous  CDF with PDF $g(u)$ on the unit interval.
 For given $S_2$, the derivative of transformation (\ref{Eq:GSED}) identifies the parent survival function $S_1$ as follows:
\begin{equation}
\label{Eq:GSEDS1}
S_1(x)=\frac{\mu_1}{\mu_2} S_2(x)g(S^e_2(x))=\frac{S_2(x)}{g(1)}g(S^e_2(x)).
\end{equation}
 Letting $x\to 0$ gives $\mu_1 =\mu_2/g(1)$.
 It is noteworthy to mention that in the case of $S_1(x)=G(S_2(x))$,
the relationship between the $\mu_2$ and $\mu_1$  can be very complicated; for example, see the PO transformation of the Weibull model (Marshall and Olkin, 1997).

\begin{Example} \rm
Let $S_2(x)=e^{-\lambda x}, x\geq 0$ and $G$ be the PO model.
Then $S^e_2$   is also exponential and
$$
S^e_1(x) = \frac{\alpha e^{-\lambda x}}{1-\bar\alpha e^{-\lambda x}}, \quad  x\geq 0 .
$$
is the extended exponential model of Marshall and Olkin (1997).
By (\ref{Eq:GSEDS1}), this is the survival function of the ED of a new distribution with the following parent survival function:
$$
S_1(x)=  \frac{\alpha^2 e^{-\lambda x}}{(1-\bar\alpha e^{-\lambda x})^2}, \quad  x\geq 0 .
$$
It is known that the inputs and outputs of the PO link model are ordered by $\alpha$ as follows: $S_2(x) \leq (\geq)\ S_1(x)$ for $\alpha \leq 1 (\geq 1)$
for all $x$.
It is easy to show that with the exponential model, $S^e_2(x) \leq  (\geq)\ S_2(x)$ for $\alpha \leq 1 (\geq 1)$.
\end{Example}

R\'enyi information divergence between two scaled survival functions on support $[0,\infty)$ is
\begin{equation}
\label{Eq:KEq-S}
K_q(S_1:S_2)  =\left \{\begin{array}{lll}
\frac{1}{q-1} \log \int_0^\infty ( S_1(x)/\mu_1)^q (S_2(x)/\mu_2)^{1-q}dx , & q >0, q\neq 1,\\ [.1in]
  K(S_1:S_2), & q=1 ,
\end{array} \right .
\end{equation}
 where   $K(S_1:S_2)=\lim_{q \uparrow 1}K_q(S_1:S_2)$ and
\begin{equation}
\label{Eq:KE-S}
 K(S_1:S_2)= \frac{1}{\mu_1} \int_0^\infty  S_1(x)\log \frac{S_1(x)}{S_2(x)}dx-\log \frac{\mu_1}{\mu_2}
\end{equation}
is the KL information divergence between two scaled survival functions.
$K_q(S_1:S_2)\geq 0$, and the inequality becomes equality if and only if $S_1(x)=S_2(x)$ almost everywhere.
The equality of the two survival functions is seen by noting that $K(f^e_1:f^e_2) = 0$ if and only if $S_1(x)/\mu_1=S_2(x)/\mu_2$ almost everywhere
and in particular, at the limit, $x\to 0$ gives $\mu_1=\mu_2$.

Divergence (\ref{Eq:KEq-S}) is in the same vein as the exponential survival entropy studied by Zografos and Nadarajah (2005)
defined for $|X|$; ($K_q(S_1:S_2)$ for distributions with support $\Re$ is defined in terms of $|X|$).
The measure (\ref{Eq:KE-S}) is  a special case of a generalized divergence defined in Asadi et el.\,(2017) in terms of the generalized logarithm.
It is the divergence version of the ED entropy studied by Asadi et el.\,(2014).

In light of the fact that
\begin{equation}
\label{Eq:KED}
K_q(S_1:S_2)  = K_q(f^e_1:f^e_2),
\end{equation}
all properties of the KL and R\'enyi divergences between two PDFs holds for  $K(S_1:S_2)$ and $K_q(S_1:S_2)$.
In particular, under  (\ref{Eq:GSED}),
\begin{eqnarray*}
&& K(S_1:S_2) =K(f_1:f_2) =\int_0^1 g(u)  \log  g(u)   du\\
&& K(S_2:S_1)=K(f_2:f_1)  =-\int_0^1   \log  g(u)   du.
 \end{eqnarray*}
Hence, the condition of Proposition  \ref{Pro:K12=K21S} for symmetry of the divergences between $f^e_1$ and $f^e_2$
holds for $K(S_1:S_2)$ under  (\ref{Eq:GSED}). Also other generalizations of  KL divergence between two PDFs extends to $K(S_1:S_2)$.

\begin{Remark} \rm
Another  KL-type divergence between two survival functions has been defined in the literature  as follows:
$$
K_s(S_1:S_2)=  \int_0^\infty S_1(x) \log  \frac{S_1(x)}{S_2(x)} d x +\mu_2-\mu_1 \geq 0.
$$
 This measure is known as the Cumulative Residual Kullback-Leibler information; see for example, Baratpour and Habibi Rad (2012) and  Park, et al.\,(2012).
Unlike, $K(S_1:S_2)$, R\'enyi version of $K_s(S_1:S_2)$ is not available and the condition of Proposition  \ref{Pro:K12=K21S}
does not hold for its symmetry under transformation (\ref{Eq:GSED}).
For this model,  $K_s(S_1:S_2)= K_s(S_2:S_1)$ if and only if
$$
K(g:g^*)-g(1)K(g^*:g)=2[1-g(1)]  +[1+g(1)]\log  g(1),
$$
where $g^*$ is the uniform PDF on $[0,1]$.
 However, we have not been able to find a link function $G$ that satisfies this condition.
\end{Remark}

\subsection{Dependence information divergence}
Let $f$ denotes the joint PDF of $(X,Y)$ with marginal PDFs $f_k, k=1,2$.
The dependence information divergence is given by
$$
K_q(f:f_1f_2) = \left \{\begin{array}{lll}
 \frac{1}{q-1}\log \int\int  f^q(x,y)f_1^{1-q}(x)f_2^{1-q}dxdy, & q >0, q\neq 1,\\ [.1in]
  K(f:f_1f_2) , & q=1,
\end{array} \right .
$$
where $K(f:f_1f_2) =\lim_{q \uparrow 1}K_q(f:f_1f_2)$ is the widely used mutual information measure of dependence between the two random variables.

The information divergence is invariant under one-to-one transformations of each variable.
The copula transformation of $(X, Y)$ is defined by $U=F_1(X), V=F_2(Y)$, and
$$
  C(u,v)= F(F^{-1}_1(u),F^{-1}_2(v)).
$$
Copula is a bivariate CDF on $[0,1]\times [0,1]$ with PDF $c(u,v)$ and uniform marginal distributions.
The relationship between $f(x,y)$ and the copula PDF is as follows:
\begin{equation}
\label{Eq:CopulaPDF}
   f(x,y) = c(F_1(x),F_2(y))f_1(x)f_2(y),
\end{equation}
where $f_1$ and $f_2$ denote the marginal densities of $X$ and $Y$, respectively.
$$
K_q(f:f_1f_2) = \left \{\begin{array}{lll}
 K_q(c:c^*), & q >0, q\neq 1,\\ [.1in]
  K(c:c^*)=-H(c) , & q=1,
\end{array} \right .
$$
where $c^*(u,v)=1$ is the PDF of independent copula $C(u,v)=uv$.
Using this property we obtain the following corollary.

\begin{Corollary} \label{Cor:Coupla}
Let $F$ be a bivariate CDF with a joint PDF  $f$, marginal PDFs $f_k, k=1,2$, and copula PDF $c$.
Then the dependence information divergence of $f$ is symmetric, $K_q(f:f_1f_2) = K_q(f_1f_2:f)$, if and only if
$$ \left \{\begin{array}{lll}
   K_q(c:c^*)=K_q(c^*:c), & q >0, q\neq 1,\\ [.1in]
   H(c)= E^*[\log c(U,V)] , & q=1 ,
\end{array} \right .
$$
where $E^*$ denotes the expectation with respect to the PDF of independent copula $c^*(u,v)=1$.
\end{Corollary}

For well known copulas, such as Gaussian and Farlie-Gumbel-Morgenstern, the mutual information $K(f:f_1f_2)$ does not
satisfy the symmetric condition of Corollary \ref{Cor:Coupla}. Hence, the symmetric R\'enyi case of $K_{1/2}(f:f_1f_2)$ remains the choice.

\section*{Acknowledgement}
The authors are thankful to Farzad Parvaresh of the Department of Electrical Engineering, University of Isfahan for kindly suggesting the idea of piecewise uniform PDF for Example \ref{Ex:Asymm}. Asadi's research was carried out in IPM Isfahan branch and was in part supported by a grant from IPM (No.\,98620215). The work of Devarajan was funded in part through the NIH/NCI Cancer Center Support Grant P30\,CA006927.  Soofi's research was  supported by the 2019 and 2020 Business Advisory Council Faculty Scholar Awards.

\appendix

\section*{Appendix}
\subsection*{Proof of Proposition \ref{Pro:K12=K21S}}
Let  $u=F_2(x)$, $g$ be a PDF and $g^*$ be the uniform PDF on $[0,1]$. Then:
\begin{eqnarray}
\label{Eq:Hq}
 K_q(f_1:f_2)=K_q(g:g^*) &=& \frac{1}{q-1} \log \int_0^1 g^q(u)du, \quad  q>0, \ q\neq 1 ;\\
\nonumber
  &=&  -H(g),  \quad  q= 1 .
\end{eqnarray}
Similarly,
\begin{eqnarray}
\label{Eq:H(1-q)}
 K_q(f_2:f_1)= K_q(g^*:g) &=& \frac{1}{q-1}  \log \int_0^1 g^{1-q}(u)du,  \quad  q>0, \ q\neq 1 ;\\
 \nonumber
 &=&  - \int_0^1 \log  g (u)du =E^*(\log g(X)), \quad  q=1 .
\end{eqnarray}
 R\'enyi entropy of order $q>0$ is defined by the negative of the integral in (\ref{Eq:Hq}).
The integral in (\ref{Eq:H(1-q)}) is $ -qH_{1-q}(g)/(q-1)$  defined for $q < 1$.

\subsection*{Proof of Proposition \ref{Pro:K12=K21F}}
The PDFs in (\ref{Eq:GLL}) are related as follows:
$$
f_1(x)=f_{2}(x)\frac{g(G^{-1}(F_2(x))+\theta)}{g(G^{-1}(F_2(x))}.
$$
\begin{enumerate}
  \item[(a)]
Letting $u=F_2(x)$ and $v=G^{-1}(x)$  we obtain the following integrals:
\begin{eqnarray*}
K(f_1:f_2) = - \int_{-\infty}^{\infty}g(v)\log g(v+\theta)dv ,\\
K(f_2:f_1) = - \int_{-\infty}^{\infty}g(v)\log g(v-\theta)dv.
\end{eqnarray*}
  \item[(b)]
  Similarly it can be shown that
\begin{eqnarray*}
\label{Eq:GLLf2}
  &&\int_{-\infty}^{\infty} f_2^{q}(x)f_1^{1-q}(x)dx    = \int_{-\infty}^{\infty} g^{1-q}(v+\theta)g^{q}(v) dv ,\\
 \label{Eq:GLLf3}
&&  \int_{-\infty}^{\infty} f_2^{q}(x)f_1^{1-q}(x)dx =  \int_{-\infty}^{\infty} g^{1-q}(v-\theta)g^{q}(v) dv.
\end{eqnarray*}
\end{enumerate}


\begin{thebibliography}{7}
\expandafter\ifx\csname natexlab\endcsname\relax\def\natexlab#1{#1}\fi

\bibitem{asadi}
Asadi, M., Ebrahimi, N., Soofi, E. S., Zohrevand, Y. (2016).  Jensen-Shannon information of the coherent system lifetime.
{\it Reliability Engineering \& System Safety}, 156, 244-255.
\bibitem{Bagdon:1997}
Bagdonavicius, V.~B., Nikulin, M.~S.  (1997). Transfer functionals and semi-parametric regression models. \emph{Biometrika}, 84, 365-378.

\bibitem{Baratpour}
Baratpour, S., \& Habibi Rad, A.  (2012).  Testing goodness-of-fit for exponential distribution based on cumulative residual entropy.
{\em   Communications in Statistics - Theory and Methods}, 41,  1387-1396.

\bibitem{Bernardo:2002}
Bernardo, J.~M. and  Rueda,  R.  (2002). Bayesian hypothesis testing: a  reference approach. {\em International Statistics Review}, 70, 351--372.

\bibitem{ch}
Chang, W-C., Li, C-L., Yang, Y., Póczos, B. (2019). Kernel Change-Point Detection with Auxiliary Deep Generative Models.
{\em Proceedings of ICLR 2019}, arXiv:1901.06077.

\bibitem{ch1}
Chen, D., Xue, Y., Gomes, C.P. (2018). End-to-End learning for the deep multivariate probit model. arXiv:1803.08591v4.

\bibitem{Cox:1972}
Cox, D.~R.  (1972).  Regression models and life tables (with Discussion). \textit{J. R. Statist. Soc. {\rm B}}, 34, 187--220.

\bibitem{er}
Eruhimov, V., Martyanov, V., Tuv, E. (2007). Change-point detection with supervised learning and feature selection.
{\em Proceedings of ICINCO 2007 - International Conference on Informatics in Control, Automation and Robotics}, 359-363.

\bibitem{gr}
Gomez-Rodriguez, M., Leskovec, J., Scholkopf, B. (2013). Modeling Information Propagation with Survival Theory. arXiv:1305.3616v1.

\bibitem{Granger:2004}
Granger, C.W.,  Maasoumi, E., Racine, J.  (2004). A dependence metric for possibly nonlinear processes. {\em Journal of Time Series Analysis}, 25, 649-669.

\bibitem{Hirschberg:1991}
Hirschberg, J., Maasoumi, E., Slottje, D.~J.  (1991). Cluster analysis and  the quality of life across countries.  {\em Journal of Econometrics},   50,  131-150.

\bibitem{doks}
Doksum, K. A., Miriam Gasko, M. (1990). On a correspondence between models in binary regression analysis and in survival analysis.
{\em International Statistical Review}, 58, 243-252.

\bibitem{kull}
Kullback, S. (1959). {\it Information Theory and Statistics}. New York: Wiley (reprinted in 1968 by Dover).

\bibitem{Lin}
 Lin, J. (1991). Divergence measures based on the Shannon entropy.  {\em IEEE Transactions on Information Theory},  37,   145-151.

\bibitem{Marshall:1997}
Marshall, A.~W.,  Olkin, I.  (1997). A new method for adding a parameter to a family of distributions with application to the exponential and Weibull families.
{\em Biometrika},  84,  641-652.

\bibitem{marti}
 Martinussen, T., Scheike, T. H. (2006). {\em Dynamic Regression Models for Survival Data}, Springer, New York.

\bibitem{Nelson:2019}
Nielsen, F. (2019). On the Kullback-Leibler divergence between location-scale densities.  arXiv:1904.10428v2.

\bibitem{Park}
 Park, S., Rao, M., \& Shin, D.~W.   (2012). On cumulative residual Kullback-Leibler information.
{\em Statistics \& Probability Letters}, 82, 2025-2032.

\bibitem{perez1}
P\'erez-Ortiz, M., Guti\'errez, P. A., Cruz-Ram\'irez, M., S\'anchez-Monedero, J., Herv\'as-Martínez, C. (2013).
Kernelizing the proportional odds model through the empirical kernel mapping.
In: Rojas I., Joya G., Gabestany J. (eds) {\em Advances in Computational Intelligence. IWANN 2013. Lecture Notes in Computer Sciencez},
vol. 7902, 270-279, Springer, Berlin.

\bibitem{perez2}
P\'erez-Ortiz, M., Guti\'errezy, P.A., Tinoz, P., Casanova-Mateox, C., Salcedo-Sanz, S. (2019).
A mixture of experts model for predicting persistent weather patterns. arXiv:1903.10012v1.

\bibitem{Ross}
 Ross, S.~M.  (1983). {\em Stochastic Processes}, John Wiley, New York.

\bibitem{Amari}
Seghouane, A-K.,  Amari, S-I. (2002). The AIC Criterion and symmetrizing the Kullback–Leibler divergence. {\em IEEE Transactions on Neural Networks}, 18, 97-106.

\bibitem{Shaked}
 Shaked, M. \& Shanthikumar, J.G.  (2007). {\em Stochastic orders}. Springer, New York.

\bibitem{soofi}
Soofi, E. S.,   Ebrahimi,  N., \& Habibullah, M. (1995). Information distinguishability with application to analysis of failure data.
{\em Journal of the American  Statistical Association},  90, 657-668.

\bibitem{Spi}
 Spirko, L.  (2017). Variable Selection and Supervised Dimension Reduction for Large-Scale Genomic Data with Censored Survival Outcomes.
\emph{Ph.D. Dissertation}. Department of Statistical Science, Temple University, Philadelphia.

\bibitem{Spi-Burns}
Spirko-Burns L., Devarajan K. (2020). Unified methods for feature selection in large-lcale Genomic studies with censored survival outcomes. {\em Bioinformatics}, 2020; vol. 36(11): 3409-3417. doi: 10.1093/bioinformatics/btaa161.


\bibitem{va}
Vargas, V.M., Guti\'errez, P.A., Herv\'as, C. (2019). {\em Deep Ordinal Classification Based on the Proportional Odds Model}. Springer International Publishing

\bibitem{wang1}
Wang, L., Madiman, M. (2014). Beyond the entropy power inequality, via rearrangements.
{\em IEEE Transactions on Information Theory}, 60(9), 5116-5137

\bibitem{wang2}
Wang, L., Zhu, D. (2019). Tackling multiple ordinal regression problems: Sparse and deep multi-task learning approaches. arXiv:1907.12508v1.

\bibitem{wu}
Wu, J., Crawford, F.W., Kim, D.A., Stafford, D., Christakis, N.A. (2018). Exposure, hazard, and survival analysis of diffusion on social networks. {\em Statistics in Mededicine}, 37, 2561–2585. doi:10.1002/sim.7658.

\bibitem{Yang:2005}
Yang, S., Prentice, R.  (2005). Semiparametric analysis of short-term and long-term hazard ratios with two-sample survival data. \emph{Biometrika},  92, 1-17.

\bibitem{yin}
Yin, X., Huang, J.G., Li, Z., Zhou, X. (2013). A survival modeling approach to biomedical search result diversification using Wikipedia.
{\em IEEE Transactions on Knowledge and Data Engineering},  25,  1201-1212.

\bibitem{Zografos:2005}
Zografos, K.,  Nadarajah, S.  (2005). Survival exponential entropies. {\em IEEE Transactions on Information Theory},  51, 1239-1246.

\bibitem{zheng}
Zheng, S., Liu, W. (2012). Functional gradient ascent for Probit regression. {\em Pattern Recognition}, 45, 4428-4437.
\end{thebibliography}

\end{document}